\documentclass{article}
\usepackage{ijcai21}

\usepackage{times}
\usepackage{soul}
\usepackage{url}
\usepackage[hidelinks]{hyperref}
\usepackage[utf8]{inputenc}
\usepackage[small]{caption}
\usepackage{graphicx}
\usepackage{amsmath}
\usepackage{amsthm}
\usepackage{booktabs}
\usepackage{algorithm}
\usepackage{amsfonts}       
\usepackage{nicefrac}       
\usepackage{microtype}      
\usepackage{multirow}
\usepackage{tabu}
\DeclareMathOperator{\trace}{tr}

\usepackage{algpseudocode}

\usepackage[caption=false]{subfig}
\usepackage{array}
\usepackage{tabularx}
\usepackage{subfig}
\usepackage{wrapfig}
\usepackage[table]{xcolor}
\usepackage{bbm}

\newtheorem{theorem}{Theorem}
\newtheorem{lemma}[theorem]{Lemma}

\title{Few-shot Learning via Dependency Maximization \\ and Instance Discriminant Analysis}

\author{
Zejiang Hou
\and
Sun-Yuan Kung
\affiliations
Princeton University\\
\emails
\{zejiangh, kung\}@princeton.edu
}

\begin{document}

\maketitle

\begin{abstract}
We study the few-shot learning (FSL) problem, where a model learns to recognize new objects with extremely few labeled training data per category. Most of previous FSL approaches resort to the meta-learning paradigm, where the model accumulates inductive bias through learning many training tasks so as to solve a new unseen few-shot task. In contrast, we propose a simple approach to exploit unlabeled data accompanying the few-shot task for improving few-shot performance. Firstly, we propose a \textit{Dependency Maximization} method based on the Hilbert-Schmidt norm of the cross-covariance operator, which maximizes the statistical dependency between the embedded feature of those unlabeled data and their label predictions, together with the supervised loss over the support set. We then use the obtained model to infer the pseudo-labels for those unlabeled data. Furthermore, we propose an \textit{Instance Discriminant Analysis} to evaluate the credibility of each pseudo-labeled example and select the most faithful ones into an augmented support set to retrain the model as in the first step. We iterate the above process until the pseudo-labels for the unlabeled data becomes stable. Following the standard transductive and semi-supervised FSL setting, our experiments show that the proposed method outperforms previous state-of-the-art methods on four widely used benchmarks, including \textit{mini}-ImageNet, \textit{tiered}-ImageNet, CUB, and CIFARFS. 
\end{abstract}

\section{Introduction}
Deep learning approaches have achieved remarkable performance on visual recognition
problems such as image classification. However, the success of deep neural
networks hinges on the availability of vast quantities of labeled training examples.
The expensive human annotation cost and the scarcity of data in some rare species
will limit their applicability to learn new concepts quickly and efficiently. In contrast,
human intelligence has the ability to quickly learn new concepts from extremely
few labeled examples, by leveraging the prior experience and integrating it with a small amount of new information. Just
as humans can efficiently learn new concept, it is desirable for the deep learning
models to learn novel classes of objects with very limited labeled examples as well. This
learning approach is referred to as the few-shot learning (FSL).

FSL has recently received substantial research interests, with a large body of work focusing on the meta-learning paradigm and episodic training strategy. In meta-learning, the model is trained on a series of episodes, with support and query examples, that simulate the generalization during testing time. After accumulating the prior experience, the trained model may have the ability to generalize to novel classes with only few labeled data. However, \cite{chen2019closer} empirically found that meta-learning may not demonstrate performance advantage, compared to the simplest baseline with a linear classifier coupled with deep feature extractor.

More recent methods start exploring transductive and semi-supervised learning for few-shot tasks, by leveraging the information from unlabeled query examples or an additional unlabeled set. Among various methods, self-training \cite{raina2007self} is one of the most straight-forward way to utilize the unlabeled data. Typically, a model trained on the support examples can be used to infer the pseudo-labels (class that has the maximum predicted probability) of the unlabeled data, and then uses these pseudo-labels along with the support set to retrain the model for predicting the query examples. However, in few-shot learning, since the model is trained with very few labeled support examples, it may not capture the data distribution of target classes in the task. Thus, the pseudo-labels may be of low quality. Including wrongly labeled examples into the training set may jeopardize the final model performance.

\paragraph{Our contributions.} We present a simple approach to exploit the unlabeled examples to improve few-shot performance. Firstly, we propose a \textit{Dependency Maximization} loss to enhance the model training, which maximizes the statistical dependence between the embedded features of unlabeled data and their softmax predictions, in conjunction with the supervised loss minimization over support set. To this end, we develop a empirical dependence measure based on the Hilbert-Schmidt norm of the cross-covariance operator. We then use the obtained model to infer the pseudo-labels for those unlabeled data, where we further propose an \textit{Instance Discriminant Analysis} to evaluate the sample from the perspective 
of feature discriminant power and select the most faithful pseudo-labels to augment the support set and retrain the model. Following the standard transductive and semi-supervised FSL, our extensive experiments show that our method compares favourably with state-of-the-art methods, not only on the widely adopted few-shot benchmarks, but on more challenging scenarios such as cross-domain FSL and higher-way testing classes. 

\section{Related Works}
We briefly review recently proposed few-shot learning approaches, focusing in more details on transductive and semi-supervised FSL. Optimization-based meta-learning methods \cite{finn2017model,antoniou2018train,rusu2018meta,sun2019meta} learn the model through a series of episodes, so that it can adapt to new tasks of novel categories with limited labeled examples. In constrast, our method does not resort to the complex meta-training; we use a feature extractor pretrained on the base classes with standard cross-entropy loss. Metric learning based methods learn to compare feature similarity based on some distance metric between support and query examples in the feature space. Examples of distance metrics include cosine similarity \cite{vinyals2016matching}, Euclidean distance \cite{snell2017prototypical,ye2020few}, relation network \cite{sung2018learning,hou2019cross}, mahalanobis distance \cite{bateni2020improved}, Earth Mover's distance \cite{zhang2020deepemd}, subspace projection distance \cite{simon2020adaptive}. In this paper, we do not utilize specialized distance metric, instead we propose a label-free dependency maximization loss for task inference. Hallucination based methods \cite{gao2018low,zhang2018metagan,li2020adversarial} utilize generative models or data augmentations to expand the support set by synthesizing new samples or features based on the given labeled data.

\paragraph{Transductive and Semi-supervised FSL.}
In practical applications, we may have unlabeled data accompanying the few-shot task, apart from the labeled support set. Transductive FSL (TFSL) methods assume that the query examples come in as a bulk and can be used as unlabeled data to faciliate the few-shot performance. To name a few, \cite{liu2018learning} utilizes label-propagation to propagate labels from labeled to unlabeled examples via a graph. \cite{rodriguez2020embedding} proposes embedding-propagation regularizer for manifold smoothing. \cite{hu2020empirical} proposes a Laplacian regularizer to encourage nearby query samples to have consistent label assignments. \cite{dhillon2019baseline} proposes to minimize the conditional entropy of the query softmax predictions. Similarly, \cite{boudiaf2020transductive} further incorporates a marginal entropy of the query softmax predictions, which helps to avoid degenerate solutions obtained when solely minimizing conditional entropy. In contrast, our method proposes to maximize the statistical dependency (DM loss) between the features and their label predictions. In semi-supervised FSL (SSFSL), the unlabeled data comes in addition to the support/query set. To name a few, \cite{li2019learning} applies self-labeling and soft-attention to the unlabeled set with finetuning on both labeled and self-labeled examples. \cite{ren2018meta} proposes a prototype refinement based on the soft assignment scores for the unlabeled examples. \cite{wang2020instance} introduces a linear regression hypothesis to select pseudo-labeled examples for classifier training. Different from these approaches, we propose a simple instance discriminant analysis, together with our DM loss, to utilize the unlabeled data for improving FSL performance.

\section{Methodology}
\subsection{Few-show Learning Formulation}
Assume we are given a labeled base dataset $\mathcal{X}_{base}=\{(\mathbf{x}_i, \mathbf{y}_i), \mathbf{y}_i\in\mathcal{Y}_{base}\}$, where $\mathcal{Y}_{base}$ denotes the set of classes (i.e. category set) for the base dataset. Few-shot learning (FSL) entails learning a model on the base dataset such that it is able to classify unseen data into a set of novel classes, given very limited labeled examples per class. Assume a novel dataset $\mathcal{X}_{novel}=\{(\mathbf{x}_i,\mathbf{y}_i),\mathbf{y}_i\in\mathcal{Y}_{novel}\}$ with completely new category set $\mathcal{Y}_{novel}$, from which we sample the few-shot tasks. We emphasize that the base and novel datasets have mutually disjoint classes of objects, i.e. $\mathcal{Y}_{base}\cap\mathcal{Y}_{novel}=\emptyset$. We follow the standard $N$-way $K$-shot task formulation. Specifically, for each few-shot task $\mathcal{T}_i$, we randomly sample $N$ classes from $\mathcal{Y}_{novel}$. We then sample $K$ labeled examples for each of $N$ classes and construct the task \textit{support set} $\mathcal{D}_{\mathcal{T}_i}^S$ with set size $|\mathcal{D}_{\mathcal{T}_i}^S|=N\times K$. Each task also has a \textit{query set} $\mathcal{D}_{\mathcal{T}_i}^Q$, which consists of $Q$ unlabeled and unseen examples for the same $N$ classes, i.e. $|\mathcal{D}_{\mathcal{T}_i}^Q|=Q\times K$. The unlabeled query set serves to evaluate the generalization performance of the model trained on base set and also adapted on the labeled support set. 

One of the fundamental challenge for FSL is the difficulty to estimate the data distribution of novel categories with only one or few labeled examples. To address this problem, a lot of recent FSL approaches resort to semi-supervised learning (SSFSL) or transductive learning (TFSL), by utilizing unlabeled examples $\mathcal{D}_{\mathcal{T}_i}^U$ of novel categories for the task at hand. In SSFSL setting, extra examples (unlabeled) apart from the support and query examples are available for the model to learn to solve the task, while TFSL assumes the model evaluates all query examples at once and utilizes those query examples as the unlabeled set. Among various semi-supervised learning methods, self-training \cite{raina2007self} is one of the state-of-the-art representatives that can be easily applied. Specifically, let $f_{\theta}: \mathcal{X}\rightarrow \mathcal{Z}\subset R^{d}$ denote the feature extractor of a deep neural network parameterized by $\theta$, where $\mathcal{Z}$ denotes the space of the feature embedding. In this work, we pretrain the feature-extractor on the labeled base dataset $\mathcal{X}_{base}$, following existing FSL approaches \cite{boudiaf2020transductive}. Given a few-shot task $\mathcal{T}_i$, self-training first learns a classifier on the labeled support set: $\underset{\phi}{\text{min}}~~\sum_{(\mathbf{x}, \mathbf{y})\in\mathcal{D}_{\mathcal{T}_i}^S}\mathcal{L}(h_{\phi}(f_{\theta}(\mathbf{x})), \mathbf{y})$ where $\mathcal{L}$ is the standard cross-entropy loss. Then, the classifier is used to infer the pseudo-labels $\hat{y}_u=h_{\phi}(f_{\theta}(\mathbf{x}_u))$ for the unlabeled examples $\mathcal{D}_{\mathcal{T}_i}^U=\{\mathbf{x}_u\}_{u=1}^U$. The pseudo-labeled examples are taken as additional labeled data for the corresponding classes and are augmented with the support examples using their pseudo-labels as true labels. Finally, the classifier is retrained using the augmented support set and evaluated on the query set.

Despite its simplicity, self-training based SSFSL/TFSL suffers from two limitations: (1) since the classifier $h_{\phi}(\cdot)$ is trained with very few labeled support examples, the pseudo-labels can be of low quality with significant label noise; (2) lack of sample selection strategy to identify and remove outliers (examples from distractor classes or untrustworthy pseudo-labeled examples for the target classes) will jeopardize the final accuracy. To tackle them, we propose a \textit{Dependency Maximization} loss to enhance the classifier training for generating pseudo labels of higher quality, and an \textit{Instance Discriminant Analysis} to evaluate the pseudo-labeled examples and select the most trustworthy ones to augment the support set.


\subsection{Dependency Maximization}
In this section, we introduce a \textit{Dependency Maximization} (DM) loss, which is differentiable and can be optimized with standard gradient descent algorithm to enhance the classifier training. While we train the classifier with the labeled support examples for inferring the unlabeled set, we propose to maximize the statistical dependence between the features of the unlabeled set and their label predictions, in conjunction with minimizing the cross-entropy loss over the support set. The DM loss can be regarded as a surrogate for the classifier's empirical risk defined over the unlabeled examples, which helps to restrict the classifier's hypothesis space and facilitates the prediction for the given unlabeled examples.

We begin by listing some notations before introducing how to characterize the dependence between features and label predictions. Let $Z$ denote the random variable associated with the embedded features of the unlabeled set, $Y$ denote the random variable associated with their softmax predictions, and $P_{Z,Y}$ be the joint distribution between these two random variables. To measure the dependence between $Z$ and $Y$, we define the cross-covariance operator based on \cite{baker1973joint}:
\begin{equation}
C_{zy} := \mathbb{E}_{zy}[(\Phi(z)-\mu_{z})\otimes(\Psi(y)-\mu_{y})]
\end{equation}
where $\Phi:\mathcal{Z}\rightarrow\mathcal{F}$ ($\Psi:\mathcal{Y}\rightarrow\mathcal{G}$) defines a kernel mapping from the space of feature embedding (space of prediction vector) to a reproducing kernel Hilbert space (RKHS) $F$ ($G$), with means defined as $\mu_z$ ($\mu_y$). $\otimes$ denotes the tensor product. A statistic that can efficiently summarize the degree of dependence between $Z$ and $Y$ is the Hilbert-Schmidt norm of the operator $C_{zy}$, which is given by the trace of $C_{zy}C_{zy}^T$. In this paper, we consider the square of the Hilbert-Schmidt norm of the cross-covariance operator, $\|C_{zy}\|_{HS}^2$, as it can detect arbitrary dependence.
\begin{theorem}[\cite{gretton2005measuring}]
\textit{Assume $F$ and $G$ are RKHSs with characteristic kernels. Then, $\|C_{zy}\|_{HS}^2=0$ if and only if Z and Y are independent.}
\end{theorem}
Characteristic kernels such as Gaussian kernel, i.e. $k(x,x')=\text{exp}\big(-\|x-x'\|_2^2/(2\sigma^2)\big)$, allows us to measure any dependence between $Z$ and $Y$. In our case, $\|C_{zy}\|_{HS}^2$ is zero only if the features and the label predictions of the unlabeled set are independent. Clearly, we aim to achieve the opposite, namely to maximize the dependence between features and predictions via maximizing the value of $\|C_{zy}\|_{HS}^2$.

To utilize the dependence measure as a loss function for classifier training, we need an empirical estimate from finite number of samples. Formally, denote the kernel functions associated with the RKHS $F$ and $G$ as $k(z,z')$ and $l(y,y')$; let $\mathbf{K},\mathbf{L}\in\mathbb{R}^{U\times U}$ denote the Gram matrices defined over the features and softmax predictions associated with the unlabeled set $\mathcal{D}_{\mathcal{T}_i}^U$, containing entries $\mathbf{K}_{i,j}=k(z_i,z_j)$ and $\mathbf{L}_{i,j}=l(y_i,y_j)$. Then, an empirical estimator of $\|C_{zy}\|_{HS}^2$ is given as:
\begin{equation}
\widehat{\|C_{zy}\|_{HS}^2} := (U-1)^{-2}\trace({\mathbf{K}\mathbf{H}\mathbf{L}\mathbf{H}})
\end{equation}
where $\mathbf{H}=\mathbf{I}_U-({1}/{U})\mathbf{1}_U\mathbf{1}_U^T$ is the centering matrix, $\mathbf{I}_U$ is an identity matrix, $\mathbf{1}_U$ is a vector with all ones, and $\trace(\cdot)$ is the matrix trace operation. We show by the following theorem that this empirical estimator converges sufficiently.
\begin{theorem}[\cite{gretton2005measuring}]
Assume $k$ and $l$ are bounded almost
everywhere by 1, and are non-negative. Then, with constants $\alpha^2>0.24$ and $C$, for $U > 1$ and all $\delta > 0$, with probability at
least $1-\delta$ for all $P_{ZY}$, we have
\begin{equation}
|\widehat{\|C_{zy}\|_{HS}^2} - \|C_{zy}\|_{HS}^2| \leq \sqrt{\dfrac{{log}(6/\delta)}{\alpha^2U}} + \dfrac{C}{U}
\end{equation}
\end{theorem}

With the empirical estimator, we can now define the overall loss function for classifier training, by integrating the empirical dependence measure defined over the unlabeled set into the supervised cross-entropy loss defined over the support set:
\begin{equation}
\resizebox{1.0\hsize}{!}{$
\underset{\mathbf{W},\mathbf{b}}{\text{min}}\underbrace{-\dfrac{1}{NK}\hspace{-0.15in}\sum_{(x,y)\in\mathcal{D}_{\mathcal{T}_i}^S}\hspace{-0.15in}\text{log}\dfrac{\text{exp}(\mathbf{W}_y^Tf_\theta(\mathbf{x})+\mathbf{b}_y)}{\text{exp}(\sum_{c=1}^N\mathbf{W}_c^Tf_\theta(\mathbf{x})+\mathbf{b}_c)}}_{\text{Cross-entropy minimization on support set}} \underbrace{- \lambda\cdot(U-1)^{-2}\trace({\mathbf{K}\mathbf{H}\mathbf{L}\mathbf{H}})}_{\text{Dependency maximization on unlabeled set}}$
}\label{DM loss}
\end{equation}
where $\mathbf{W},\mathbf{b}$ denote the weight and bias of the softmax linear classifier $h_{\phi}$, and the label prediction is given by $\hat{\mathbf{y}}=h_{\phi}(\mathbf{z})=\text{softmax}(\mathbf{W}^T\mathbf{z}+\mathbf{b})=\text{softmax}(\mathbf{W}^Tf_\theta(\mathbf{x})+\mathbf{b})$. 

Objective (\ref{DM loss}) can be optimized for each test task via standard gradient descent (GD) algorithm w.r.t. $\mathbf{W}$ and $\mathbf{b}$. Specifically, during classifier training, the pretrained feature extractor $f_{\theta}$ is frozen. $\mathbf{W}$ and $\mathbf{b}$ are initialized based on the class prototypes computed over the support set: $\mathbf{W}^0=[2\boldsymbol{\mu}_1,...,2\boldsymbol{\mu}_N]\in\mathbb{R}^{d\times N}$ and $\mathbf{b}^0=[-\|\boldsymbol{\mu}_1\|_2^2,...,-\|\boldsymbol{\mu}_N\|_2^2]\in\mathbb{R}^{N\times 1}$. Then, the weight and bias parameters will be updated via GD using both support and unlabeled samples of the few-shot task without mini-batch sampling.

\subsection{Instance Discriminant Analysis}


\begin{table}[t]
\caption{The feature discriminant power (measured by the normalized $\psi$ in Eq.\eqref{DI}) can serve as the surrogate for the pseudo-labels' accuracy over the unlabeled set, thus can be used as the criterion to evaluate the credibility of the pseudo-labeled examples. ``Random'': randomly guessing the labels for unlabeled data. ``w/o DM'': inferred pseudo-labels from a classifier trained with cross-entropy only. ``w/ DM'': inferred pseudo-labels from a classifier trained with Eq.\eqref{DM loss}.}
\vspace{-0.1in}
\small
\centering
\begin{tabular}{c c c c c}\toprule
    \multirow{2}{*}{Labeling} & \multicolumn{2}{c}{\textit{mini}-ImageNet} & \multicolumn{2}{c}{\textit{tiered}-ImageNet} \\
    & $\psi$ & Acc.(\%) & $\psi$ & Acc.(\%) \\ \midrule\midrule
    Random & 0.14 & 20.00 & 0.13 & 20.00  \\
    Pseudo (w/o DM) & 0.55 & 57.73 & 0.61 & 68.29  \\
    Pseudo (w/ DM) & 0.68 & 75.80 & 0.74 & 82.43 \\
    Groundtruth & 1.00 & 100.00 & 1.00 & 100.00 \\
    \bottomrule
\end{tabular}
\label{tab:DI_vs_Acc}
\vspace{-0.1in}
\end{table}

After training the classifier with our proposed DM loss in Eq.\eqref{DM loss}, we can now predict the labels $\hat{y}_u$ for the unlabeled examples in $\mathcal{D}_{\mathcal{T}_i}^U$ as their pseudo-labels. In this section, we present an \textit{Instance Discriminant Analysis} (IDA) to evaluate the quality of these pseudo-labeled examples and select the most trustworthy ones into the augmented support set.

IDA is essentially a sample selection or outlier removal algorithm, that aims to remove a subset of training data sample \textit{a priori}, and train the classifier only with the remaining subset of data. To this end, we introduce a hypothesis that the feature discriminant power computed on the embedded features and pseudo-labels can be used as a surrogate for the pseudo-labels' accuracy for the unlabeled set (cf. Table \ref{tab:DI_vs_Acc}). The rationale behind this is that a wrongly labeled example would be detrimental to the overall data separability of the unlabeled set, causing low feature discriminant power; while a correctly labeled example would facilitate the data separability, improving the feature discriminant power. 

In this paper, we adopt Fisher's discriminant analysis as the basis of our quality measure. We evaluate the quality of each pseudo-labeled instance by computing its contribution to the overall data separability based on the \textit{Fishers Criterion}. Formally, denote the set of pseudo-labeled data as $\{(\mathbf{x}_u,\hat{y}_u)| \mathbf{x}_u\in\mathcal{D}_{\mathcal{T}_i}^U\}$. Let $f(\mathbf{x}_u)$ denote the embedded feature of instance $u$ (for notation simplicity, we omit $\theta$ for the feature extractor). we define the \textit{scatter matrix} and the \textit{between-class scatter matrix}, respectively, as $\bar{\mathbf{S}}=\sum_{u=1}^{U}(f(\mathbf{x}_u)-\boldsymbol{\mu})(f(\mathbf{x}_u)-\boldsymbol{\mu})^T$ and
$\mathbf{S}_B=\sum_{c\in\{1,...,N\}}M_c(\boldsymbol{\mu}_c-\boldsymbol{\mu})(\boldsymbol{\mu}_c-\boldsymbol{\mu})^T$, where $\boldsymbol{\mu}$ is the mean of all embedded features associated with the pseudo-labeled set, $M_c$ is the number of instances belonging to class $c$, $\boldsymbol{\mu}_c$ is the mean of embedded features belonging to class $c$, and $N$ is the number classes in the given few-shot task. Then, the \textit{Fishers Criterion} ($\psi$) is defined as the ratio of the between-class scatter matrix to the scatter matrix:
\begin{equation}
\psi := \trace\{\bar{\mathbf{S}}^{-1}\mathbf{S}_B\}\label{DI}
\end{equation}
where $\trace(\cdot)$ denotes the matrix trace operation. To explain more, the eigen-vectors of matrix $\bar{\mathbf{S}}^{-1}\mathbf{S}_B$ composes the optimal space that maximises the between-class separability while minimising the within-class variability. The Fishers Criterion, calculated as the summation of the corresponding eigen-values, can be regarded as a measure of the overall data separability.

Next, we can evaluate the credibility of each pseudo-labeled instance $(\mathbf{x}_u,\hat{y}_u)$ by measuring its contribution to the overall discriminant power, i.e. to measure the difference of Fishers Criterion value when the instance is present and the instance is removed while keeping everything else constant. Precisely, the influence of removing a specific instance on $\psi$ is referred to as the \textit{Instance Discriminant Analysis} (IDA):
\begin{equation}
d\psi_u := \trace\{\bar{\mathbf{S}}^{-1}\mathbf{S}_B\} - \trace\{[\bar{\mathbf{S}}\neg u]^{-1}\mathbf{S}_B\neg u\}\label{IDA}
\end{equation}
where $\bar{\mathbf{S}}\neg u$ and $\mathbf{S}_B\neg u$ are derived from the remaining data after removing instance $u$. $d\psi_u$ captures the reduction in the feature discriminant power caused by removing instance $u$, and it can be used as a metric for our sample selection process. Larger $d\psi_u$ indicates that the instance has greater (positive) impact to the data separability, thus its pseudo-label is more trustworthy and the instance should be selected to the augmented support set. We sort the pseudo-labeled examples in the descending order of their $d\psi_u$ value, and only select the top-ranking examples.

On the other hand, the exact computation of $d\psi_u$ can be expensive, since it requires multiple matrix inverse. In order to perform our IDA-based sample evaluation and selection more efficiently, we provide the following theorem as an approximation of the $d\psi_u$, which can be computed without any matrix operations, with only inner-product and scaler operations. The proof is provided in the appendix.
\begin{theorem}
Instance Discriminant Analysis (IDA) $d\psi_u$ of sample $u$ is upper-bounded by:
\begin{equation}
\resizebox{1.02\hsize}{!}{$
d\psi_u\leq\dfrac{\delta f(\mathbf{x}_u)^Tf(\mathbf{x}_u)}{\rho(f(\mathbf{x}_u)^Tf(\mathbf{x}_u)-\rho)}+
\dfrac{H_{4,1/2}(\nu_u+f(\mathbf{x}_u)^Tf(\mathbf{x}_u))}{\rho(M_u-1)}+
\dfrac{f(\mathbf{x}_u)^Tf(\mathbf{x}_u)(\nu_u+f(\mathbf{x}_u)^Tf(\mathbf{x}_u))}{\rho(f(\mathbf{x}_u)^Tf(\mathbf{x}_u)-\rho)(M_u-1)}$
}
\end{equation}
where $\delta=\sum_{c\in\{1,...,N\}}M_c\boldsymbol{\mu}_c^T\boldsymbol{\mu}_c$; $\rho>0$ is the ridge parameter; $M_u$ is the number of examples sharing the same pseudo label as $\mathbf{x}_u$ in the dataset (including $\mathbf{x}_u$); $H_{4,1/2}=\sum_{k=1}^4k^{-1/2}$ is the generalized harmonic number; 
$M_c$ is number of examples that have pseudo label equal to class $c$; $\boldsymbol{\mu}_c$ is the mean of class $c$; $\boldsymbol{\mu}_u$ is the mean of class that example $\mathbf{x}_u$ belongs to; and
$\nu_u =M_u[(\boldsymbol{\mu}_u^T\boldsymbol{\mu}_u)^2-4(\boldsymbol{\mu}_u^T\boldsymbol{\mu}_u)(\boldsymbol{\mu}_u^Tf(\boldsymbol{x}_u))+2(f(\boldsymbol{x}_u)^Tf(\boldsymbol{x}_u))(\boldsymbol{\mu}_u^T\boldsymbol{\mu}_u)+2(\boldsymbol{\mu}_u^Tf(\boldsymbol{x}_u))^2]^{1/2}$.
\end{theorem}

In practice, we iteratively select the most trustworthy pseudo-labeled examples based on their IDA values to augment the support set. Specifically, the classifier is first trained with the initial support examples using Eq.\eqref{DM loss}. Then, it can be used to infer the pseudo-labels and we employ the IDA measure to select the most faithful ones into the support set. The expanded support set will be used to update the classifier based on Eq.\eqref{DM loss} again. We iterate the above process to progressively enhance the classifier until the predicted pseudo-labels for the unlabeled set becomes stable, as summarized in Algorithm 1.


\begin{algorithm}[t]
    \small
    \centering
    \caption{Few-shot Learning with Dependency Maximization and Instance Discriminant Analysis}
    \begin{algorithmic}[1]
        \State \textbf{Require} Support set $D_{\mathcal{T}_i^{S}}=\{\mathbf{x}_n,y_n\}_{n=1}^{NK}$; Unlabeled data $D_{\mathcal{T}_i}^U=\{\mathbf{x}_u\}_{u=1}^{{U}}$; pretrained feature extractor $f_{\theta}(\cdot)$ on base set.
        \State \textbf{Initialize} augmented support set $(X_s,y_s)=\{\mathbf{x}_n,y_n\}_{n=1}^{NK}$ .
        \While{pseudo-labels are not stablized}
        \State Train a classifier $h_{\phi}(\cdot)$ on $(X_s,y_s)$ and $D_{\mathcal{T}_i}^U$ using Eq.\eqref{DM loss}.
        \State Infer pseudo-labels for $\{\mathbf{x}_u\}_{u=1}^{{U}}$ and obtain $\{\mathbf{x}_u,\hat{y}_u\}_{u=1}^{{U}}$.
        \State Compute IDA for each pseudo-labeled instance using Eq.\eqref{IDA}.
        \State Rank $\{\mathbf{x}_u,\hat{y}_u\}_{u=1}^{{U}}$ based on their IDA value $d\psi_u$.
        \State Select the most trustworthy subset $(X_{sub},y_{sub})$ from $\{\mathbf{x}_u,\hat{y}_u\}_{u=1}^{{U}}$, and merge them into $(X_s,y_s)$.
        \EndWhile
        \State \textbf{Return} Augmented support set $(X_s,y_s)$.
    \end{algorithmic}
\end{algorithm}

\begingroup
\setlength{\tabcolsep}{6pt} 
\begin{table*}[t]
\caption{Comparison of testing accuracy with previous state-of-the-art methods on four few-shot benchmark datasets. ``In.'' and ``Tran." denotes inductive and transductive FSL, respectively. ``Semi.'' denotes semi-supervised FSL. ``-'' denotes the results are not provided by the corresponding method. Methods with ``$\dagger$" use WRN28-10 as the backbone network.}
\vspace{-0.1in}
\small
\centering
\begin{tabular}{c c c c c c c c c c}\toprule
    \multirow{2}{*}{Method} & \multirow{2}{*}{Setting} & \multicolumn{2}{c}{\textbf{\textit{mini}-ImageNet}} & \multicolumn{2}{c}{\textbf{CUB}} & \multicolumn{2}{c}{\textbf{CIFARFS}} & \multicolumn{2}{c}{\textbf{\textit{tiered}-ImageNet}} \\
    & & 1-shot & 5-shot & 1-shot & 5-shot & 1-shot & 5-shot & 1-shot & 5-shot \\\midrule\midrule
    DSN \cite{simon2020adaptive} & \multirow{3}{*}{In.} & 62.64 & 78.83 & - & - & 72.30 & 85.10 & 66.22 & 82.79 \\
    FEAT \cite{ye2020few} & & 66.78 & 82.05 & - & - & - & - & 70.80 & 84.79 \\
    DeepEMD \cite{zhang2020deepemd} & & 65.91 & 82.41 & 75.65 & 88.69 & - & - & 71.16 & 86.03 \\
    \midrule
    TransFinetune$^\dagger$ \cite{dhillon2019baseline} & \multirow{7}{*}{Tran.} & 65.73 & 78.40 & - & - & 76.58 & 85.79 & 73.34 & 85.50 \\
    LaplacianShot$^\dagger$ \cite{ziko2020laplacian} & & 74.86 & 84.13 & 80.96 & 88.68 & - & - & 80.18 & 87.56 \\
    TIM$^\dagger$ \cite{boudiaf2020transductive} & & 77.80 & 87.40 & 82.20 & 90.80 & - & - & 82.10 & 89.80 \\
    SIB$^\dagger$ \cite{hu2020empirical} & & 70.00 & 79.20 & - & - & 80.00 & 85.30 & - & - \\
    EPNet$^\dagger$ \cite{rodriguez2020embedding} & & 70.74 & 84.34 & 87.75 & 94.03 & - & - & 78.50 & 88.36 \\
    ICI+LR \cite{wang2020instance} & & 66.80 & 79.26 & 88.06 & 92.53 & 73.97 & 84.13 & 80.79 & 87.92 \\
    BD-CSPN$^\dagger$ \cite{liu2020prototype} & & 70.31 & 81.89 & 87.45 & 91.74 & & & 78.74 & 86.92 \\
    \midrule
    LST \cite{li2019learning} & \multirow{4}{*}{Semi.} & 70.10 & 78.70 & - & - & - & - & 77.70 & 85.20 \\
    ICA+MSP$^\dagger$ \cite{lichtenstein2020tafssl} & & 80.11 & 85.78 & - & - & - & - & 86.00 & 89.39 \\
    EPNet$^\dagger$ \cite{rodriguez2020embedding} & & 79.22 & 88.05	& - & - & - & - & 83.69 & 89.34 \\
    ICI+LR \cite{wang2020instance} & & 71.41 & 81.12 & 91.11 & 92.98 & 78.07 & 84.76 & 85.44 & 89.12\\
    \midrule
    \multirow{2}{*}{{Ours}} & Tran. & 79.17 & 87.02 & 92.43 & 94.77 & 79.52 & 86.16 & 84.10 & 89.61 \\
    & Semi. & \textbf{82.46} & \textbf{88.17} & \textbf{93.51} & \textbf{95.44} & \textbf{82.16} & \textbf{87.26} & \textbf{87.12} & \textbf{90.54} \\
    \bottomrule
\end{tabular}
\label{tab:FSL_benchmarks}
\vspace{-0.1in}
\end{table*}
\endgroup

\section{Experiments}
\subsection{Setup}
We evaluate on four widely used few-shot benchmark datasets: \textbf{\textit{mini}-ImageNet} \cite{vinyals2016matching} consists of 100 classes, and we follow the split of base/novel classes as \cite{ravi2016optimization}; \textbf{\textit{tiered}-ImageNet} contains 608 classes and we follow the spilt as \cite{ren2018meta}; \textbf{CUB} \cite{wah2011caltech} is a fine-grained classification dataset, containing 200 classes and we follow the split as \cite{chen2019closer}; \textbf{CIFARFS} is a low-resolution few-shot dataset, containing 100 classes and we follow the split as \cite{wang2020instance}.

Throughout the experiments, the hyperparameters of DM loss and IDA algorithm are kept fixed. Specifically, we use Guassian kernel with bandwith $\sigma=0.5$ for the DM loss, and weight $\lambda$ in Eq.\eqref{DM loss} is set to 0.01. For the IDA algorithm, we select at most 5 samples per class at each iteration, until the pseudo-labeling process becomes stable. For training the softmax classifier, we use ADAM optimizer with $10^{-4}$ learning rate and run 1000 iterations for each task. Unless otherwise specified, we use WRN28-10 \cite{zagoruyko2016wide} as our main backbone for feature extraction, as it has been widely used by previous works. Training of the backbone network follows the same training procedure (without episodic training) as \cite{boudiaf2020transductive} on base classes for all datasets. The models are trained for 90 epochs, with initial learning rate 0.1, divided by 10 at epochs 1/2 and 2/3, and batchsize 128. We employ standard data augmentation, including random crop, color jittering, and random horizontal flipping. All input images are of size $84\times84$. To evaluate the testing performance, we randomly sample 10,000 few-shot tasks from the novel classes and report the averaged accuracy.

\subsection{Benchmark results}
Table \ref{tab:FSL_benchmarks} evaluates our method on the four
benchmarks, under both transductive and semi-supervised setting.
\paragraph{Transductive FSL.}
In TFSL (denoted as Tran.), we have access to the query examples in the inference stage, thus we take the query set as the unlabeled set and utilize our proposed DM and IDA algorithms (i.e. no additional unlabeled set, but using the query set as the unlabeled set in Trans.) As shown in Table \ref{tab:FSL_benchmarks}, our proposed method compares favourably with recently proposed TFSL approaches across all datasets, especially in the 1-shot setting where the labeled support data is extremely limited.

\paragraph{Semi-supervised FSL.}
We follow the SSFSL (denoted as Semi.) setup in \cite{wang2020instance,rodriguez2020embedding,lichtenstein2020tafssl}, where each testing task has an additional unlabeled set consisting of unlabeled examples from the classes in the support set. In this paper, we use 50 unlabeled examples per class in both 1-shot and 5-shot scenarios. Compared with other SSFSL methods which usually use 100 unlabeled examples per class, our method shows competitive or better accuracy across all benchmark datasets. Moreover, comparing our SSFSL results versus our TFSL results, we can see that the additional unlabeled examples indeed helps to improve the final accuracy.

\paragraph{Cross-domain FSL.}
\cite{chen2019closer} recently showed that many of meta-learning algorithms perform no better than the simplest finetuning baseline when there exists a domain-shift between the base dataset for training and the novel dataset for testing. We also evaluate our method in this challenging scenario, where we train the backbone network on \textit{mini}-ImageNet while testing it on the few-shot tasks from CUB. As shown in Table \ref{tab:cross_domain}, our method compares favourably with previous meta-learning and transductive FSL methods, suggesting the potention of applicability to real-world problems.

\begin{table}[t]
\caption{Results of testing accuracy for cross-domain FSL scenario. For a fair comparison, we use the same ResNet18 backbone as compared methods. $^\S$: denote transductive FSL methods.}
\vspace{-0.1in}
\small
\centering
\begin{tabular}{c c c}\toprule
    \multirow{2}{*}{Method} & \multicolumn{2}{c}{\textit{mini}-ImageNet$\rightarrow$ CUB}  \\
     & 1-shot & 5-shot \\\midrule\midrule
     MAML \cite{finn2017model} & - & 51.34 \\
     ProtoNet \cite{snell2017prototypical} & - & 62.02 \\
     RelationNet \cite{sung2018learning} & - & 57.71 \\
     Finetuning \cite{chen2019closer} & 48.56 & 65.57 \\
     LaplacianShot $^\S$ \cite{hu2020empirical} & \textbf{55.46} & 66.33 \\
     Ours (Tran.) $^\S$ & \textbf{55.79} & \textbf{71.01} \\
     \bottomrule
\end{tabular}
\label{tab:cross_domain}
\vspace{-0.1in}
\end{table}

\paragraph{Higher-way testing scenario.}
We evaluate our method on more challenging 10-way and 20-way few-shot scenarios. 
As shown in Table \ref{tab:higher_way}, compared with previous meta-learning or transductive methods, our proposed method still achieves the highest accuracy for testing tasks with higher number of ways.

\begingroup
\setlength{\tabcolsep}{4pt} 
\begin{table}[t]
\caption{Results of testing accuracy for higher-way scenario on the \textit{mini}-ImageNet. For a fair comparison, all methods are based on the WRN28-10 backbone. $^\S$: denote transductive FSL methods.}
\vspace{-0.1in}
\small
\centering
\begin{tabular}{c c c c c}\toprule
    \multirow{2}{*}{Method} & \multicolumn{2}{c}{10-way} & \multicolumn{2}{c}{20-way} \\
     & 1-shot & 5-shot & 1-shot & 5-shot \\\midrule\midrule
    Baseline++ \cite{chen2019closer} & 40.43 &	56.89 &	26.92 &	42.80 \\
    LEO \cite{rusu2018meta} & 45.26 &	64.36 &	31.42 &	50.48 \\
    MetaOpt \cite{lee2019meta} & 44.83 &	64.49 &	31.50 &	51.25 \\
    S2M2$_R$ \cite{mangla2020charting} & 50.40 &	70.93 &	36.50 &	58.36 \\
    EPNet $^\S$ \cite{rodriguez2020embedding} & 53.70 & 72.17 & 38.55 & 59.01 \\
    BD-CSPN $^\S$ \cite{liu2020prototype} & 51.58 &	69.35 &	36.00 &	55.23 \\
    Ours (Tran.) $^\S$ & \textbf{60.05} & \textbf{75.93} & \textbf{41.47} & \textbf{61.91}\\
    \bottomrule
\end{tabular}
\label{tab:higher_way}
\vspace{-0.1in}
\end{table}
\endgroup

\subsection{Ablation study}
We investigate the effectiveness of various components proposed in our method, namely the \textit{Dependency Maximization} loss and the \textit{Instance Discriminant Analysis} via ablation study.

\begin{figure}[t]
\centering
\includegraphics[scale=0.224]{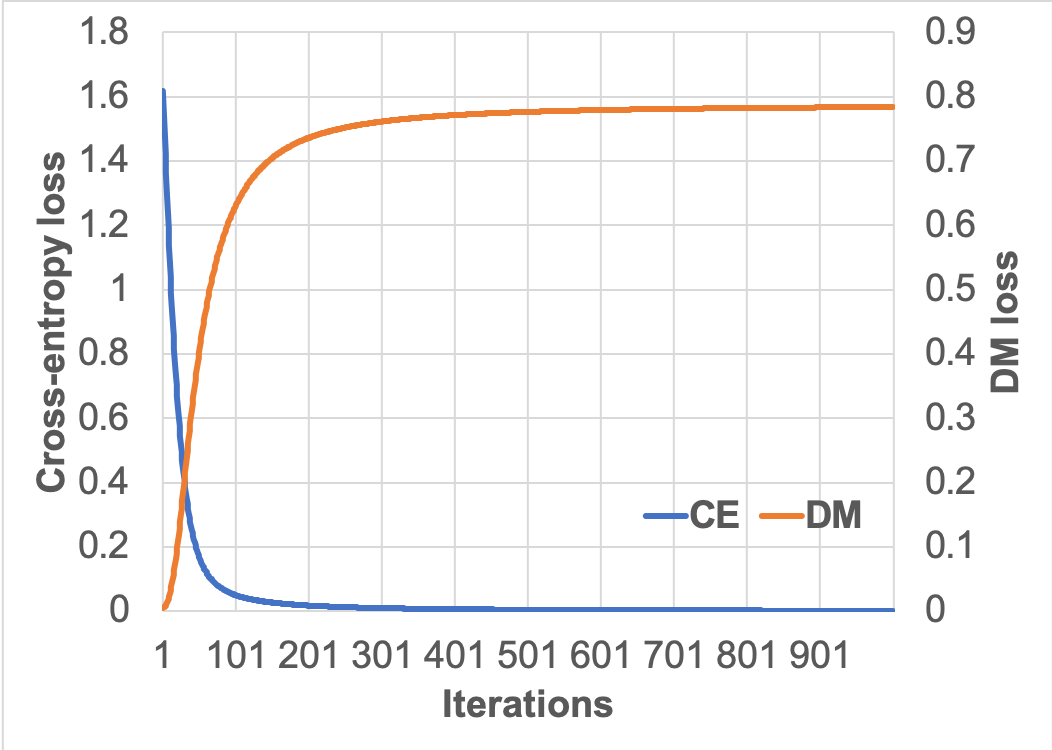}
\includegraphics[scale=0.224]{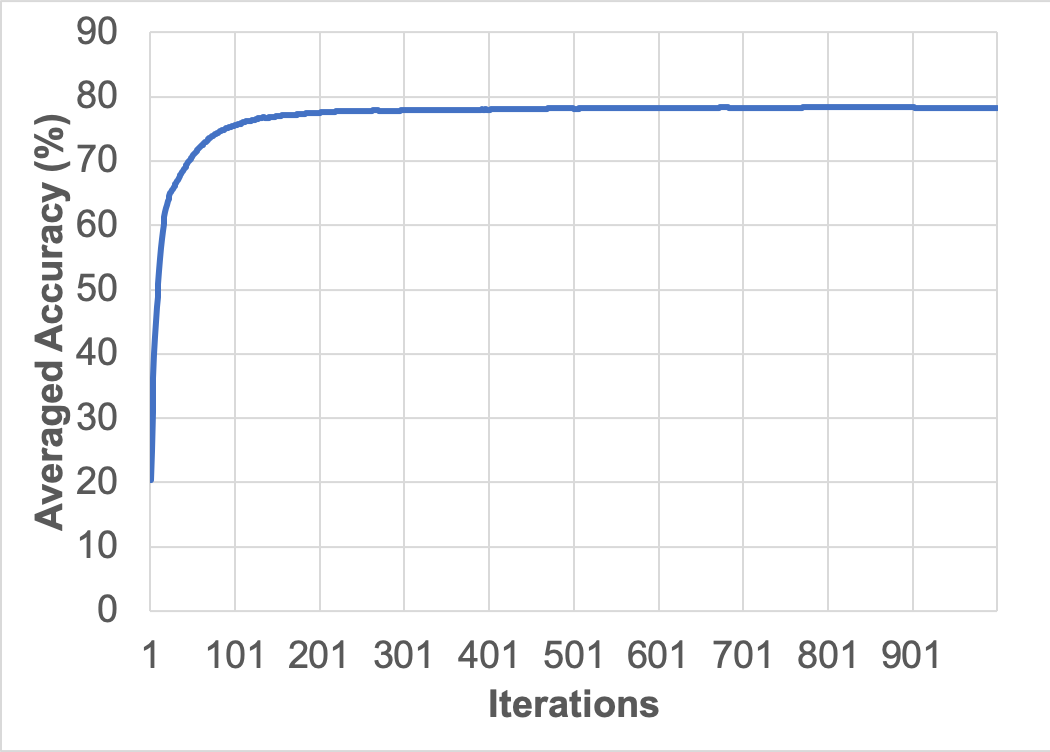}
\caption{Convergence plot while we use Eq.\eqref{DM loss} to train the classifer on \textit{mini}-ImageNet 1-shot tasks. Left: cross-entropy loss and DM loss versus iterations. Right: averaged accuracy versus iterations.}
\label{fig:convergence}
\end{figure}

\paragraph{Effectiveness of DM.} To validate the effectiveness of DM loss, we compare it with several recently proposed label-free loss functions utilizing unsupervised information in query data for transductive FSL. Results are reported in Table \ref{tab:ablation_DM}. \cite{dhillon2019baseline} proposes to minimize the conditional entropy of the label predictions over query data; \cite{boudiaf2020transductive} proposes to maximize the weighted mutual information over query data. Nevertheless, we can observe that incorporating our DM loss consistently outperform other types of transductive learning on both \textit{mini}-ImageNet and \textit{tiered}-ImageNet. This suggests that maximizing the dependency between query feature and the label predictions can effectively improve the generalization performance. Furthermore, Figure \ref{fig:convergence} shows the convergence plot for our DM method on 1-shot \textit{mini}-ImageNet tasks. One can see that during training, DM value increases monotonically at each iteration and converges well.

\begingroup
\setlength{\tabcolsep}{1pt} 
\begin{table}[t]
\caption{Ablation study on the effect of proposed MD loss. Results are reported for WRN28-10 as the backbone for different methods.}
\vspace{-0.1in}
\small
\centering
\begin{tabular}{c c c c c}\toprule
    \multirow{2}{*}{Loss} & \multicolumn{2}{c}{\textit{mini}-ImageNet} & \multicolumn{2}{c}{\textit{tiered}-ImageNet} \\
     & 1-shot & 5-shot & 1-shot & 5-shot \\\midrule\midrule
    Baseline & 57.73 & 78.17 & 68.29 & 85.31 \\
    Cond. Ent. \cite{dhillon2019baseline} & 65.73 & 78.40 & 73.34 & 85.50 \\
    Mul. Info. \cite{boudiaf2020transductive} & 71.54 & 83.92 & 77.02 & 87.57 \\
    DM (Ours) & \textbf{75.80} & \textbf{85.26} & \textbf{82.42} & \textbf{89.12} \\
    \bottomrule
\end{tabular}
\label{tab:ablation_DM}
\vspace{-0.1in}
\end{table}
\endgroup

\paragraph{Effectiveness of IDA.} To further validate the effectiveness of IDA, we compare it with other metrics for evaluating the credibility and selecting the pseudo-labeled examples under the transductive/semi-supervised FSL setting in Table \ref{tab:ablation_IDA}. A naive strategy is to randomly select some pseudo-labeled examples into the augmented support set, denoted as \textit{rand.}. Another strategy is to select high-confidence examples (denoted as \textit{confid.}), i.e. retaining pseudo labels whose largest class probability given by the classifier fall above certain threshold \cite{li2019learning}. One can also leverage the nearest-neighbour strategy (denoted as \textit{nn.}) to select the examples based their distance to the prototype of each class in the feature space. The last one we compare to is ICI \cite{wang2020instance}, which selects pseudo-labels based on a linear regression hypothesis. Here, we assume 15 unlabeled examples for each class and select 5 examples per class by different metrics to retrain the classifier on \textit{mini}-ImageNet. As shown, IDA outperforms other metrics in all settings, suggesting that IDA can select more faithful pseudo-labeled examples.

\begin{table}[t]
\small
\caption{Comparing IDA to other metrics for selecting the pseudo-labeled examples for self-training. Results are reported on \textit{mini}-ImageNet with ResNet12 as the backbone network.}
\vspace{-0.1in}
\centering
\begin{tabular}{c c c c c}\toprule
    \multirow{2}{*}{Metric} & \multicolumn{2}{c}{Transductive} & \multicolumn{2}{c}{Semi-supervised}  \\
    & 1-shot & 5-shot & 1-shot & 5-shot \\\midrule\midrule
    Baseline & 56.06 & 75.43 & 56.06 & 75.43 \\
    rand. &    59.01 &	76.38 &	59.46 &	76.58 \\
    nn. &    63.24 &	77.63 &	63.10 &	77.75 \\
    confid. &    63.29 &	77.92 &	63.57 &	77.71 \\
    ICI &    65.32 &	78.30 &	64.60 &	77.96 \\
    IDA (Ours) &    \textbf{67.17} &	\textbf{80.00} &	\textbf{67.36} &	\textbf{80.18} \\
    \bottomrule
\end{tabular}
\label{tab:ablation_IDA}
\vspace{-0.1in}
\end{table}

\vspace{-0.15in}
\section{Conclusion}
\vspace{-0.05in}
Few-shot learning is a fundamental problem in modern AI research. In this paper, we propose a simple approach to exploit unlabeled data to improve the few-shot performance. We propose a dependency maximization loss based on the Hilbert-Schmidt norm of the cross-covariance operator, which maximizes the statistical dependency between the features of unlabeled data and their label predictions. The obtained model can be used to infer the pseudo-labels for the unlabeled data. We further propose a instance discriminant analysis to evaluate the quality of each pseudo-labeled example and only select the most faithful ones to augment the support set. Extensive experiments show that our method compares favourably with state-of-the-art methods on standard few-shot benchmarks, as well as on higher-way testing tasks and cross-domain FSL. In future work, we will focus on providing a more theoretical ground for dependency maximization and discriminant-based sample selection. Moreover, we aim to generalize our method to more domains and applications beyond classification.
\vspace{-0.15in}

\small
\bibliographystyle{named}
\bibliography{ijcai21}

\section{Appendix}
Before we provide the proof for Theorem 3, we list two useful lemmas that are used repeatedly in the following.
\begin{lemma}[\cite{merikoski1994bounds}]
The non-increasingly ordered singular values of a matrix $\mathbf{M}$ obey $0\leq\sigma_i\leq\dfrac{\|M\|_F}{\sqrt{i}}$, where $\|\cdot\|_F$ denotes the matrix Frobenius norm.
\end{lemma}

\begin{lemma}[\cite{von1937some}]
Let $\sigma_i(M)$ and $\sigma_i(N)$ be the non-increasingly ordered singular values of matrices $\mathbf{M},\mathbf{N}\in\mathbb{R}^{a\times b}$. Then, $\trace\{\mathbf{M}\mathbf{N}^T\}\leq\sum_i^r\sigma_i(\mathbf{M})\sigma_i(\mathbf{N})$, where $r={min}(a,b)$.
\end{lemma}

\paragraph{Proof of Theorem 3}
\begin{proof}
The Fishers Criterion can be rewritten as $\psi=\trace\{\bar{\mathbf{S}}^{-1}\mathbf{S}_B\}$, where $\bar{\mathbf{S}}=\mathbf{F}\mathbf{F}^T$ ($\mathbf{F}$ is the matrix containing all features of the unlabeled set, arranged in columns) and $\mathbf{S}_B=\sum_{c=1}^NM_c\boldsymbol{\mu}_c\boldsymbol{\mu}_c^T=\sum_{c=1}^N\mathbf{S}_{c}$ ($\boldsymbol{\mu}_c$ is the mean feature vector of class c).
For notation clarity and simplicity, we assume that all data are centered and that data mean does not change after only one sample is removed. This is justifiable when the number of unlabeled data is sufficiently large, which is the case we consider here.

Suppose the removed instance has pseudo-label belonging to class $u$.
After removing the instance $f(\mathbf{x}_u)$, the two scatter matrices becomes: $\bar{\mathbf{S}}'=\bar{\mathbf{S}}-f(\mathbf{x}_u)f(\mathbf{x}_u)^T$ and $\mathbf{S}_B'=\mathbf{S}_B+\mathbf{S}_u'-\mathbf{S}_u=\mathbf{S}_B+\mathbf{E}_B$, where $\mathbf{S}_u'=(M_u-1)\boldsymbol{\mu}_u'\boldsymbol{\mu}_u'^T$ and $\boldsymbol{\mu}_u'=(\boldsymbol{\mu}_uM_u-f(\boldsymbol{\mu}_u))/(M_u-1)$. Then, we can rewrite:
\begin{equation}
\resizebox{1.02\hsize}{!}{$
\mathbf{E}_B=\dfrac{M_u\boldsymbol{\mu}_u\boldsymbol{\mu}_u^T-M_u\boldsymbol{\mu}_uf(\mathbf{x}_u)^T-M_uf(\mathbf{x}_u)\boldsymbol{\mu}_u^T+f(\mathbf{x}_u)f(\mathbf{x}_u)^T}{M_u-1}
$}
\end{equation}
We can then define the IDA as:
\begin{align}
& d\psi_u=\trace\{\bar{\mathbf{S}}^{-1}\mathbf{S}_B-\bar{\mathbf{S}}'^{-1}\mathbf{S}_B'\}\\\nonumber
& =\trace\{\bar{\mathbf{S}}^{-1}\mathbf{S}_B-(\bar{\mathbf{S}}-f(\mathbf{x}_u)f(\mathbf{x}_u)^T)^{-1}(\mathbf{S}_B+\mathbf{E}_B)\}
\end{align}
The latter term can be reformulated by the Woodbury identity \cite{horn2012matrix}:
\begin{align}
&(\bar{\mathbf{S}}-f(\mathbf{x}_u)f(\mathbf{x}_u)^T)^{-1}(\mathbf{S}_B+\mathbf{E}_B)\\\nonumber
&=(\bar{\mathbf{S}}^{-1}+\dfrac{\bar{\mathbf{S}}^{-1}f(\mathbf{x}_u)f(\mathbf{x}_u)^T\bar{\mathbf{S}}^{-1}}{1-f(\mathbf{x}_u)^T\bar{\mathbf{S}}^{-1}f(\mathbf{x}_u)})(\mathbf{S}_B+\mathbf{E}_B)
\end{align}
Substitute this term into the above IDA equation, we have:
\begin{align}
\resizebox{1.02\hsize}{!}{$
d\psi_u=\trace\{\dfrac{\bar{\mathbf{S}}^{-1}f(\mathbf{x}_u)f(\mathbf{x}_u)^T\bar{\mathbf{S}}^{-1}\mathbf{S}_B}{f(\mathbf{x}_u)^T\bar{\mathbf{S}}^{-1}f(\mathbf{x}_u)-1}+
\bar{\mathbf{S}}^{-1}\tilde{\mathbf{E}}_B+
\dfrac{\bar{\mathbf{S}}^{-1}f(\mathbf{x}_u)f(\mathbf{x}_u)^T\bar{\mathbf{S}}^{-1}\mathbf{E}_B}{f(\mathbf{x}_u)^T\bar{\mathbf{S}}^{-1}f(\mathbf{x}_u)-1}\} 
$}
\end{align}
where $\tilde{\mathbf{E}}_B=-{\mathbf{E}}_B$. To upper-bound $d\psi_u$, we derive an upper-bound for the three terms respectively, given that trace operation is additive.

Upper-bound for $\trace\{\dfrac{\bar{\mathbf{S}}^{-1}f(\mathbf{x}_u)f(\mathbf{x}_u)^T\bar{\mathbf{S}}^{-1}\mathbf{S}_B}{f(\mathbf{x}_u)^T\bar{\mathbf{S}}^{-1}f(\mathbf{x}_u)-1}\}$: From Lemma 2, we have:
\begin{align}
& \trace\{\dfrac{\bar{\mathbf{S}}^{-1}f(\mathbf{x}_u)f(\mathbf{x}_u)^T\bar{\mathbf{S}}^{-1}\mathbf{S}_B}{f(\mathbf{x}_u)^T\bar{\mathbf{S}}^{-1}f(\mathbf{x}_u)-1}\} \\\nonumber
& \leq\dfrac{\sum_i\sigma_i(\bar{\mathbf{S}}^{-1}\mathbf{S}_B\bar{\mathbf{S}}^{-1})\sigma_i(f(\mathbf{x}_u)f(\mathbf{x}_u)^T)}{f(\mathbf{x}_u)^T\bar{\mathbf{S}}^{-1}f(\mathbf{x}_u)-1}\\\nonumber
& \leq\dfrac{f(\mathbf{x}_u)^Tf(\mathbf{x}_u)\sigma_1(\bar{\mathbf{S}}^{-1}\mathbf{S}_B\bar{\mathbf{S}}^{-1})}{f(\mathbf{x}_u)^T\bar{\mathbf{S}}^{-1}f(\mathbf{x}_u)-1}
\end{align}
where $\sigma_1(\cdot)$ denotes the largest singular value. Given that the largest singular value is actually the spectral norm, based on the norm submultiplicative, we have:
\begin{equation}
\sigma_1(\bar{\mathbf{S}}^{-1}\mathbf{S}_B\bar{\mathbf{S}}^{-1})\leq\|\bar{\mathbf{S}}^{-1}\|_2^2\|\mathbf{S}_B\|_2
\end{equation}
For the first norm, $\|\bar{\mathbf{S}}^{-1}\|_2=1/\sigma_{min}(\bar{\mathbf{S}})$. Typically, $\bar{\mathbf{S}}$ is regularized by a ridge parameter $\rho>0$, i.e. $\bar{\mathbf{S}}+\rho\mathbf{I}$, it can be said that $\sigma_{min}(\bar{\mathbf{S}})>\rho$, so that $\|\bar{\mathbf{S}}^{-1}\|_2<1/\rho$. For the second norm, $\|\mathbf{S}_B\|_2=\|\sum_{c=1}^NM_c\boldsymbol{\mu}_c\boldsymbol{\mu}_c^T\|_2\leq\sum_{c=1}^NM_c\|\boldsymbol{\mu}_c\boldsymbol{\mu}_c^T\|_2=\sum_{c=1}^NM_c\boldsymbol{\mu}_c^T\boldsymbol{\mu}_c=\delta$. It follows that $\sigma_1(\bar{\mathbf{S}}^{-1}\mathbf{S}_B\bar{\mathbf{S}}^{-1})\leq\delta/\rho^2$. Finally, based on the von Neumann \cite{von1937some} property, $f(\mathbf{x}_u)^T\bar{\mathbf{S}}^{-1}f(\mathbf{x}_u)-1=\trace\{f(\mathbf{x}_u)^T\bar{\mathbf{S}}^{-1}f(\mathbf{x}_u)\}-1=C{\sigma_1(\bar{\mathbf{S}}^{-1})f(\mathbf{x}_u)^Tf(\mathbf{x}_u)}-1$, where $C\in[-1,1]$. Hence, for simplicity, we use the following approximation: $f(\mathbf{x}_u)^T\bar{\mathbf{S}}^{-1}f(\mathbf{x}_u)-1\approx f(\mathbf{x}_u)^Tf(\mathbf{x}_u)/\rho-1$. Then, we can derive the upper-bound for $\trace\{\dfrac{\bar{\mathbf{S}}^{-1}f(\mathbf{x}_u)f(\mathbf{x}_u)^T\bar{\mathbf{S}}^{-1}\mathbf{S}_B}{f(\mathbf{x}_u)^T\bar{\mathbf{S}}^{-1}f(\mathbf{x}_u)-1}\}$ as:
\begin{equation}
\trace\{\dfrac{\bar{\mathbf{S}}^{-1}f(\mathbf{x}_u)f(\mathbf{x}_u)^T\bar{\mathbf{S}}^{-1}\mathbf{S}_B}{f(\mathbf{x}_u)^T\bar{\mathbf{S}}^{-1}f(\mathbf{x}_u)-1}\}\leq\dfrac{\delta f(\mathbf{x}_u)^Tf(\mathbf{x}_u)}{\rho(f(\mathbf{x}_u)^Tf(\mathbf{x}_u)-\rho)}
\end{equation}

Upper-bound for $
\trace\{\bar{\mathbf{S}}^{-1}\tilde{\mathbf{E}}_B\}$: From Lemma 2, we have:
\begin{equation}
\trace\{\bar{\mathbf{S}}^{-1}\tilde{\mathbf{E}}_B\}\leq\sum_{i=1}^4\sigma_i(\bar{\mathbf{S}}^{-1})\sigma_i(\tilde{\mathbf{E}}_B)
\end{equation}
since $\text{rank}(\tilde{\mathbf{E}}_B)\leq4$ \cite{horn2012matrix}. Then, with Lemma 1, we have $\sigma_i(\tilde{\mathbf{E}}_B)\leq\dfrac{\|\tilde{\mathbf{E}}_B\|_F}{\sqrt{i}}=\dfrac{\|{\mathbf{E}}_B\|_F}{\sqrt{i}}$. By substituting the definition of ${\mathbf{E}}_B$ and using the triangular inequality, we have:
\begin{equation}
\resizebox{1.02\hsize}{!}{$
\sigma_i(\tilde{\mathbf{E}}_B)\leq\dfrac{\|M_u\boldsymbol{\mu}_u\boldsymbol{\mu}_u^T-M_u\boldsymbol{\mu}_uf(\mathbf{x}_u)^T-M_uf(\mathbf{x}_u)\boldsymbol{\mu}_u^T\|_F+\|f(\mathbf{x}_u)f(\mathbf{x}_u)^T\|_F}{(M_u-1)\sqrt{i}}
$}
\end{equation}
Based on the property that $\|M\|_F^2=\trace(M^TM)$:
\begin{equation}
\sigma_i(\tilde{\mathbf{E}}_B)\leq\dfrac{\nu_u+f(\mathbf{x}_u)^Tf(\mathbf{x}_u)}{(M_u-1)\sqrt{i}}
\end{equation}
where the definition of $\nu_u$ is listed in Theorem 3 of our paper. With the bound on $\sigma_1(\bar{\mathbf{S}}^{-1})<1/\rho$, we can derive the upper-bound for $\trace\{\bar{\mathbf{S}}^{-1}\tilde{\mathbf{E}}_B\}$ as:
\begin{equation}
\resizebox{1.02\hsize}{!}{$
\trace\{\bar{\mathbf{S}}^{-1}\tilde{\mathbf{E}}_B\}\leq\sum_{i=1}^4\dfrac{\nu_u+f(\mathbf{x}_u)^Tf(\mathbf{x}_u)}{\rho(M_u-1)\sqrt{i}}\leq\dfrac{H_{4,1/2}(\nu_u+f(\mathbf{x}_u)^Tf(\mathbf{x}_u))}{\rho(M_u-1)}
$}
\end{equation}

Upper-bound for $\trace\{\dfrac{\bar{\mathbf{S}}^{-1}f(\mathbf{x}_u)f(\mathbf{x}_u)^T\bar{\mathbf{S}}^{-1}\mathbf{E}_B}{f(\mathbf{x}_u)^T\bar{\mathbf{S}}^{-1}f(\mathbf{x}_u)-1}\}$: With similar derivation as in the first term, we have:
\begin{align}
& \trace\{\dfrac{\bar{\mathbf{S}}^{-1}f(\mathbf{x}_u)f(\mathbf{x}_u)^T\bar{\mathbf{S}}^{-1}\mathbf{E}_B}{f(\mathbf{x}_u)^T\bar{\mathbf{S}}^{-1}f(\mathbf{x}_u)-1}\} \\\nonumber
& \leq\dfrac{f(\mathbf{x}_u)^Tf(\mathbf{x}_u)\sigma_1(\bar{\mathbf{S}}^{-1}\mathbf{E}_B\bar{\mathbf{S}}^{-1})}{f(\mathbf{x}_u)^Tf(\mathbf{x}_u)/\rho-1}
\end{align}
Again, based on the norm submultiplicative, $\sigma_1(\bar{\mathbf{S}}^{-1}\mathbf{E}_B\bar{\mathbf{S}}^{-1})\leq\|\bar{\mathbf{S}}^{-1}\|_2^2\|\mathbf{E}_B\|_2$. From the derivation in the second term, we readily get $\|\mathbf{E}_B\|_2=\sigma_1(\|\mathbf{E}_B\|_2)\leq\|\mathbf{E}_B\|_F\leq\dfrac{\nu_u+f(\mathbf{x}_u)^Tf(\mathbf{x}_u)}{(M_u-1)}$. Using the upper-bound for $\|\bar{\mathbf{S}}^{-1}\|_2$, we can obtain the bound $\sigma_1(\bar{\mathbf{S}}^{-1}\mathbf{E}_B\bar{\mathbf{S}}^{-1})\leq\|\mathbf{E}_B\|_F\leq\dfrac{\nu_u+f(\mathbf{x}_u)^Tf(\mathbf{x}_u)}{(M_u-1)\rho^2}$. Finally, we can derive the upper-bound for the third term:
\begin{equation}
\resizebox{1.02\hsize}{!}{$
\trace\{\dfrac{\bar{\mathbf{S}}^{-1}f(\mathbf{x}_u)f(\mathbf{x}_u)^T\bar{\mathbf{S}}^{-1}\mathbf{E}_B}{f(\mathbf{x}_u)^T\bar{\mathbf{S}}^{-1}f(\mathbf{x}_u)-1}\}\leq\dfrac{f(\mathbf{x}_u)^Tf(\mathbf{x}_u)(\nu_u+f(\mathbf{x}_u)^Tf(\mathbf{x}_u))}{\rho(f(\mathbf{x}_u)^Tf(\mathbf{x}_u)-\rho)(M_u-1)}
$}
\end{equation}
Finally, we can conclude the upper-bound for $d\psi_u$ by combining the upper-bounds for three additive terms together.
\end{proof}


\end{document}